%% file: template.tex
\documentclass{Interspeech}

\usepackage{multirow}
\usepackage{multicol}

\interspeechcameraready

\title{Towards Pretraining Robust ASR Foundation Model\\ with Acoustic-Aware Data Augmentation}

\author[affiliation={1}]{Dancheng}{Liu}
\author[affiliation={1}]{Amir}{Nassereldine}
\author[affiliation={1}]{Chenhui}{Xu}
\author[affiliation={1}]{Jinjun}{Xiong}

\affiliation{}{University at Buffalo}{USA}
\email{\{dliu37,amirnass,cxu26,jinjun\}@buffalo.edu}

\usepackage{comment}

\begin{document}

\maketitle

\input{0_abstract}
\input{1_intro_v2}
\input{2_related}
\input{3_methodology_v3}

\input{5_linguistic}

\input{4_acoustic}
\input{6_conclusion}

\bibliographystyle{IEEEtran}
\bibliography{mybib}

\end{document}

%% file: 0_abstract.tex
\begin{abstract}
Whisper’s robust performance in automatic speech recognition (ASR) is often attributed to its massive 680k-hour training set—an impractical scale for most researchers. In this work, we examine how linguistic and acoustic diversity in training data affect the robustness of the ASR model and reveal that transcription generalization is primarily driven by acoustic variation rather than linguistic richness. 
We find that targeted acoustic augmentation methods could significantly improve the generalization ability of ASR models, reducing word-error rates by up to 19.24\% on unseen datasets when training on the 960-hour Librispeech dataset. These findings highlight strategic acoustically focused data augmentation as a promising alternative to massive datasets for building robust ASR models, offering a potential solution to future foundation ASR models when massive human speech data is lacking. 
\end{abstract}

%% file: 1_intro_v2.tex
\section{Introduction}
Automatic speech recognition (ASR) technology empowers numerous real-world applications from interactive voice assistants to educational and medical systems. Among existing ASR models, Whisper \cite{radford2022robustspeechrecognitionlargescale} stands out for its robust performance across diverse audio inputs, particularly accented and children's speech, where most ASR models face significant challenges \cite{liu2024automaticscreeningchildrenspeech}. 

A key factor behind Whisper’s robustness is the diversity in its training set. 
However, when acquiring such diverse data, the reported 680k hours of audio incur significant costs \cite{radford2022robustspeechrecognitionlargescale}. Even through web scraping, such massive amounts of data remain out of reach for most researchers. This challenge underscores the need for more accessible strategies to assemble large-scale training datasets. Some potential strategies to enhance diversity in the training set are through synthetic data generation \cite{fazel21_interspeech,hilmes24_syndata4genai} and data augmentation techniques on real data \cite{park19e_interspeech,kim21c_interspeech}, where the breadth of large-scale corpora can be approximated when truly diverse data is scarce. Conceptually, speech diversity decomposes into two key aspects: linguistic features (the content/transcription) and acoustic features (the speaker’s characteristics). While speech synthesis may broaden linguistic diversity with varied transcriptions, data augmentation 
may expand acoustic diversity by altering speaker-related acoustic variations in a small training dataset. However, to effectively leverage either of these alternatives and train robust foundation ASR models, we must first address the following question: 

\textbf{\textit{How is the ASR model's robustness linked to the linguistic and acoustic diversity of its training data differently?}}

This paper seeks to answer this crucial question. Through analysis and empirical results, we show that acoustic diversity plays a more critical role in the robustness of foundation ASR models than linguistic diversity. Additionally, we propose three key arguments regarding the development of end-to-end foundation ASR models. First, despite Whisper’s autoregressive nature, its underlying functional behavior differs from that of a large language model. The correctness of transcription (which could be perceived as the generation quality in language models) benefits less from textual diversity in the training corpus. A pre-trained ASR model exhibits no generalization ability without exposure to acoustic variations beyond its synthetic data distribution. Second, some of the de facto speech augmentation strategies, such as SpecAugment \cite{park19e_interspeech}, which are used by foundation models including Whisper-v2 \cite{radford2022robustspeechrecognitionlargescale}, contribute primarily to linguistic diversity rather than acoustic diversity. As a result, it is less effective than techniques that introduce more direct acoustic variations to the training dataset. Third, proper data augmentations through enhancing acoustic diversity in small training datasets can significantly boost the generalization ability of pre-trained ASR models, leading to accuracy improvements in out-of-distribution speech.
Our experimental results show that
    training a foundation ASR model 
    with
    \emph{only 960 hours} from Librispeech \cite{librispeech} dataset can achieve
    up to \textbf{19.24\%} WER reduction using acoustic-centric data augmentation.
    %

\noindent \textbf{Contributions.} In summary, our contributions are as follows:
\begin{enumerate}
    \item We discuss the difference between linguistic and acoustic variations in human speech and their connections to the robustness of foundation ASR models.
    \item We show that ASR models benefit mainly from the acoustic diversity in the training data than
    the linguistic diversity.
    \item We demonstrate 
    that acoustic-centric data augmentation strategies can significantly improve 
    the robustness of ASR models, especially when the training data is not massive.

\end{enumerate}


%% file: 2_related.tex
\section{Related Works}
\subsection{Speech Data Augmentation Techniques}
\begin{figure*}
    \centering
    \includegraphics[width=0.98\textwidth]{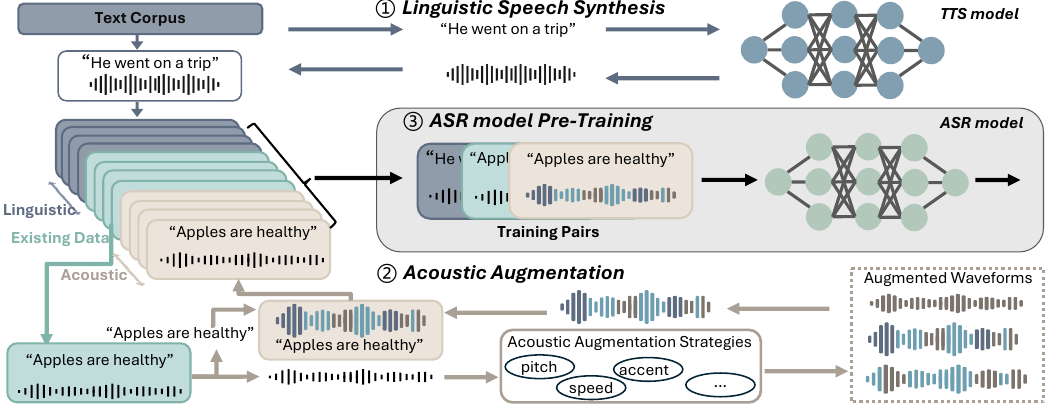}    \caption{High-level pipeline of pre-training and ASR model with synthetic data and with data augmentation strategies on the real data.}
    \vspace{-3mm}
    \label{fig:illustration}
\end{figure*}
Speech data augmentation has been a well-established field in the research community. From the early works leveraging simple transformations such as adding noise \cite{hannun2014deepspeechscalingendtoend}, changing playback speed \cite{ko15_interspeech}, applying reverberation \cite{ko2017reverb}, and concatenation \cite{concatenate}, to more modern approaches like SpecAugment \cite{park19e_interspeech}, SpecMix \cite{kim21c_interspeech}, and SpecAugment++ \cite{wang21d_interspeech}, data augmentation has been consistently proven 
effective in enhancing the robustness and generalization of speech models. 

Among these existing data augmentation techniques, SpecAugment \cite{park19e_interspeech} is an elegant and simple approach favored by current ASR model pre-training, including Whisper-v2 \cite{radford2022robustspeechrecognitionlargescale}.

The most valuable components of SpecAugment, according to their ablation studies, are frequency and time masking, which randomly removes rows and columns from the spectrogram. SpecMix \cite{kim21c_interspeech} combines the idea of SpecAumgnet \cite{park19e_interspeech} and Mixup \cite{zhang2017mixup} to create time and frequency masks using other speech samples instead of zeroing them out. Later in Section \ref{sec:acoustic}, we will show that SpecAugment \cite{park19e_interspeech} is actually a suboptimal augmentation strategy due to its indirect relationship with acoustic variations. Data augmentations such as direct perturbations on vowels and SpecMix \cite{kim21c_interspeech} provide better generalization towards out-of-distribution data for pre-trained ASR foundation models.

There exist other works proposing adaptive augmentation strategies, including SapAugment \cite{sapaugment}, G-Augment \cite{gaugment}, and improvement over SapAugment with Progressive Scheduling \cite{lu2024sampleadaptivedataaugmentation}. 
These advanced methods bring better performance to the ASR models but are less frequently used in the pre-training stage due to their complexity. Since these methods are fundamentally the same as the non-adaptive methods, and our focus is not to study the best specific data augmentation method, we omit the discussion of those works in later sections. 

\subsection{Usage of Synthetic Data in ASR Training}
While augmentation methods focus on manipulating existing real-world recordings, generating entirely synthetic speech data has also gained substantial traction in recent years. Recent advances in text-to-speech (TTS) systems (e.g., Tacotron \cite{tacotron}, WaveGAN \cite{donahue2019wavegan}, Kokoro \cite{hexgrad_2025}), have made it possible to produce highly natural and intelligible synthetic utterances. 

However, as of now, using synthetic data in the speech model's training is a largely unexplored direction. Preliminary results have been reported on improvements over accented speech recognition \cite{do2024improvingaccentedspeechrecognition,masson24_syndata4genai}, low-resource languages \cite{casanova23_interspeech}, and dysarthric speech recognition \cite{leung2024trainingdataaugmentationdysarthric}. SynthASR \cite{fazel21_interspeech} is one of the first works to investigate training with a large amount of synthetic data, and work from \cite{hilmes24_syndata4genai} is the first to investigate the effect of pre-training with purely synthetic data. However, all existing works that use synthetic data as part of training sources either have the test distribution known from either the real speech data or the synthetic data. Many works such as \cite{hilmes24_syndata4genai} use the same distribution for training and testing data (for example, Librispeech to train the TTS and then Librispeech for evaluation of the ASR). These practices are not practical for pre-training foundation ASR models by most researchers, as the training data for TTS is also limited. In this work, we investigate the effect of synthetic data on a general speech distribution that is not seen during the training process of both the TTS and ASR models.

%% file: 3_methodology_v3.tex
\section{Analysis of Data Augmentation from a Human Speech Perspective}
\label{preliminary}

As mentioned earlier, human speech is conceptually a combination of how people speak (acoustics) and what they speak (linguistics). In this section, we provide a high-level discussion of how these linguistic and acoustic features affect the diversity of the ASR model's training set, which in turn links to the robustness of the trained ASR model. 

Figure~\ref{fig:illustration} shows a high-level pipeline of the ASR model's pre-training stage. The existing data are the available data one has at the start of the pre-training stage. Each data sample contains the waveform of speech, which is a representation of the combination of acoustics, linguistics, and other confounding factors such as noises, and a paired transcription with the waveform. Without loss of generalizability, we could say that the existing data samples collectively form an acoustic space and a linguistic space, which the ASR model will learn from.

The diversity of linguistic features can be increased with synthetically generated speech data, which is shown by Part 1 of Figure~\ref{fig:illustration}. When linguistic features are disentangled from acoustic features, they represent the norms of natural language and human habits of using natural language in daily life. Thus, a diverse text-based corpus (such as from Wikipedia or from daily dialogs \cite{dai2022dialoginpainting}) will be a good approximation of the linguistic space in a massive pre-training set of an ASR model. When augmenting the linguistic space of a limited dataset, a TTS model converts the text corpus into the desired audio and text tuples, and such synthetically generated data will be used in the pre-training stage of the ASR model. 

Additionally, acoustic features can be augmented by manipulating the waveforms or their respective spectrograms to broaden the acoustic diversity of the training data. This is illustrated in Part 2 in Figure~\ref{fig:illustration}, where artificially mutating the waveforms of existing training data based on acoustic augmentation strategies generates multiple waveforms aligned with the exact transcription but exhibiting distinct acoustic characteristics.
For instance, when a speaker produces the vowel ``\textit{i},'' the waveform contains a few dominant sinusoids at different frequencies, which correspond to specific frequency bins in the spectrogram. Augmentation of such speech to simulate different accents can be performed by varying the intensities on those frequency bins. In Section \ref{sec:acoustic}, we will provide further details on the acoustic augmentation strategies, which increase the diversity of the training set.

Through synthetic data and data augmentation techniques, we may get a better approximation of the acoustic and linguistic space of a truly diverse training set from the limited data. As a direct result of the augmentations, the dataset in the final pre-training stage (Part 3 in Figure~\ref{fig:illustration}) may represent a broader acoustic or linguistic space compared to the existing real data. 
Yet, which space is more responsible for the robustness of the ASR model? In Section \ref{sec:linguistic} and \ref{sec:acoustic}, we will show through experiments that diversity in acoustic features is the key to robustness in the foundation ASR model, and acoustics-based data augmentation strategies will significantly increase the performance of the pre-trained model on out-of-distribution speech data.

%% file: 5_linguistic.tex
\section{Pre-training with Linguistic Diversity}
\label{sec:linguistic}

Results from prior works using synthetically generated data seem to indicate that synthetic data could lead to model robustness \cite{fazel21_interspeech,hilmes24_syndata4genai}. Indeed, synthetic data could increase the diversity of transcription in the training data, which will fill in the gaps of missing inter-word dependencies in natural languages when the training data is scarce. 
However, it shall be noted that since synthetic data generated from a TTS model have very limited acoustic variations, the synthetic acoustic space is much smaller than the actual acoustic space from human speech. 
When training with a very small acoustic space,  the ASR model is not able to generalize to any audio beyond synthetically generated audio as it has never seen the different acoustic patterns that do not exist in the training data. To validate this claim, we perform the following experiment. 

\noindent\textbf{Experiment Setup.} We train an ASR model model using a purely synthetic dataset. The text corpus is obtained from Wiki-Dialog \cite{dai2022dialoginpainting} and contains around 11 million conversations. We use each sentence in a conversation as an utterance and use Kokoro-v0.19 \cite{hexgrad_2025,kaneko2022istftnetfastlightweightmelspectrogram,li2023styletts2humanleveltexttospeech}, a lightweight TTS model to convert it to speech. The ``af'' voice pack is used, and all other hyperparameters remain unchanged. Due to resource constraints, throughout the paper, we will use the Whisper-base architecture and its training hyperparameters (besides batch size changed to 64) whenever experiments are conducted. All experiments are conducted on a server with 4 A6000 GPUs.

We train our model until convergence, which surprisingly takes less than 2\% of the training corpus.
We then evaluate the pre-trained ASR model on both synthetic data and real-world data. Our experiments show that the ASR model pre-trained with synthetic data is able to easily transcribe the synthetic data with a WER close to 0. In contrast, it is unable to transcribe real-world data, resulting in a WER greater than 100\%. We notice that the outputs from the ASR model on non-synthetic audio inputs are non-coherent repetitions of single letters or words. Such a phenomenon supports our claim that pre-training on linguistic diversity does not bring generalization. Since human speech and synthetically generated speech have different acoustic features, the human speech spectrograms presented to the ASR during inference are not recognized. 

%% file: 4_acoustic.tex
\section{Pre-training with Acoustic Diversity}
\label{sec:acoustic}

In Section \ref{sec:linguistic}, empirical results show that pre-training with only synthetic diversity limits the model's ability to generalize beyond the synthetic distribution.
In this section, we focus on acoustic data augmentation techniques and investigate their effects on the robustness of the pre-trained ASR models. 

\subsection{Why is SpecAugment not Optimal?}
Before diving into our augmentation strategies, it is crucial to first examine the shortcomings of existing ones—particularly the widely used SpecAugment \cite{park19e_interspeech}. In their paper, the authors of SpecMix \cite{kim21c_interspeech} suggest that SpecAugment’s deficiency arises from the loss of salient audio information due to zero masking. While this claim is technically correct, we argue that it does not capture the core reason behind SpecAugment’s shortcomings.

Following the characterization of human speech in Section \ref{preliminary} and empirical results obtained from Section \ref{sec:linguistic}, we argue that the time and frequency masking in SpecAugment only indirectly augments linguistic features. By masking out segments of the spectrogram, it forces the ASR model to “guess” missing words, creating an (augmented) mapping between partial sounds and complete transcriptions. However, since human speakers do not naturally make partial sounds, this approach merely addresses the requirements of linguistic diversity in an implicit way, while overlooking the need for acoustic diversity. Consequently, improvements reported by SpecAugment \cite{park19e_interspeech} are partly explained by the fact that the test dataset shares the same distribution as the training set, making large-scale acoustic variations less necessary for the task.

\subsection{Acoustic-Centric Data Augmentation}
\label{sec:details}
A simplified decomposition of human acoustic variations is to identify the prosody and articulation of the speech, where prosody describes the non-verbal aspects of the speech, such as pitch, amplitude, and duration, and articulation describes the verbal variations including accents. This decomposition, while not perfect, hints at several data augmentation techniques that could be useful to improve the generalization during pre-training. Due to resource limitations, we are not able to conduct extensive experiments and investigate many combinations of acoustic augmentations \footnote{Each training takes up to 15 days on the available GPUs we have.}. Results reported in Section \ref{sec:results} are augmented with pitch, amplitude, duration, and with simplified modeling for accents. Nonetheless, our results show significant improvement in the robustness of the ASR model compared to other techniques, and we hope that our preliminary results can shed light on the potential future directions of foundation ASR model pre-training. 


On the prosody side, pitch and amplitude are two straightforward augmentations. The training set, Librispeech-960h \cite{librispeech}, only contains adult speech. Thus, using Liborsa \cite{mcfee_2024_4923181}, we randomly change the pitch according to Table \ref{pitch_table}. The Probability column represents the percentage of samples that have their pitches changed within the lower and upper bound. Each sample will only be changed up to once.
Notable augmentations here are the first row simulating elderly people by lowering the pitch, and the last row simulating children's speech by uplifting the pitch. In addition, we randomly multiply the amplitude of the speech sample by  0.5 to 1.5. 

\input{table/main_table}
\input{table/pitch}

\begin{figure}[!h]
    \centering
    \includegraphics[width=0.98\linewidth]{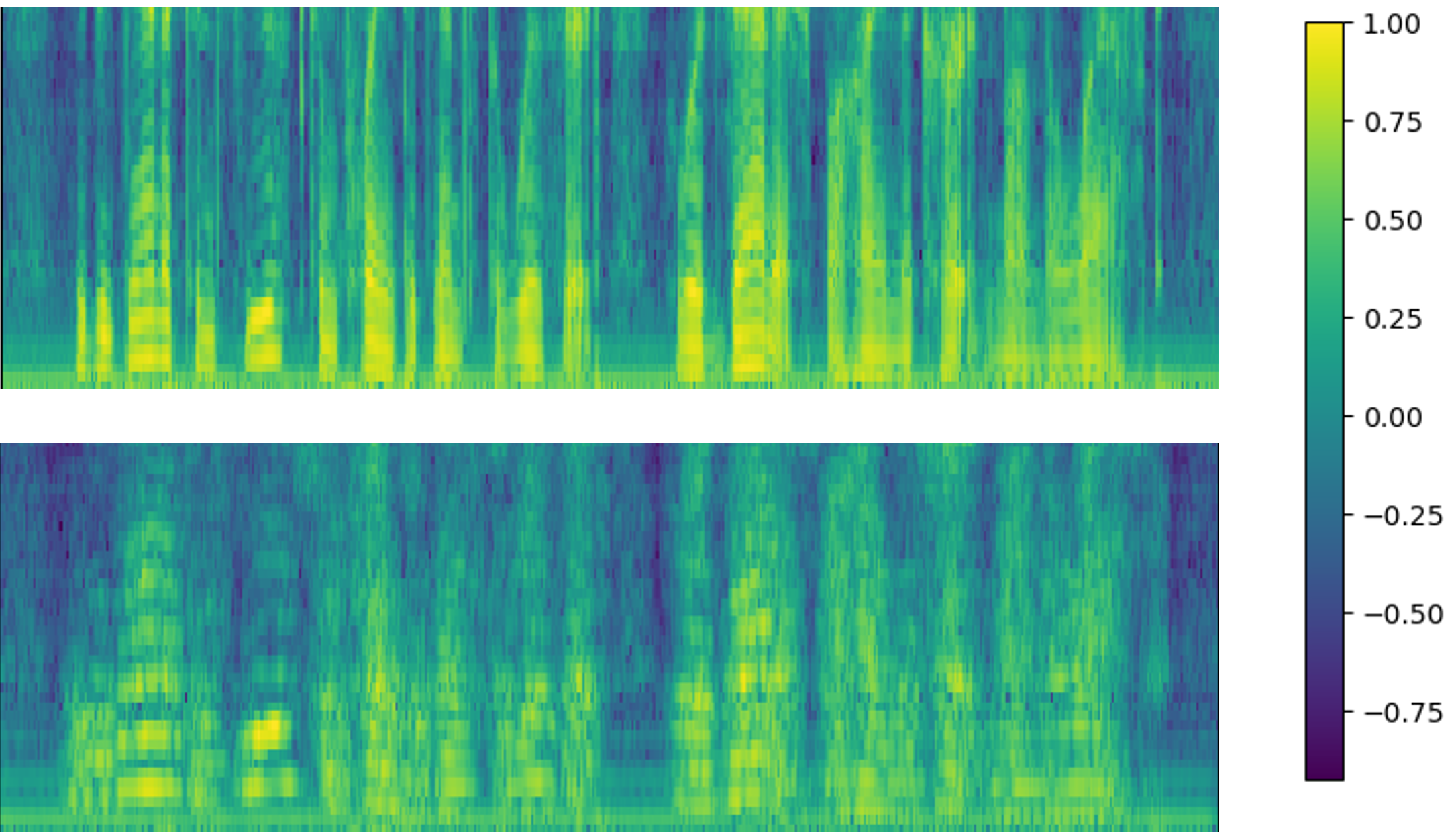}
    \caption{Normalized log mel spectrogram before (top) and after (bottom) acoustic augmentation.}
    \label{fig:spec}
    \vspace{-3mm}
\end{figure}
We tentatively combine the augmentation of duration and accent. Due to the nature of the English language where consonant variations are much less than vowels \cite{Cao2017/12}, and due to the much smaller intensity of the consonant pronunciations, the augmentation is only applied to the vowel pronunciations. The steps of augmentation are as follows: \textbf{1.} normalize the magnitude of the spectrogram between 0 and 1; \textbf{2.} identify the columns representing vowel pronunciations by a threshold (0.3 in the implementation) and group those columns by adjacency (each group then represents one complete vowel pronunciation); \textbf{3.} randomly change the duration of vowel pronunciations by repeating or removing some columns of vowel pronunciations; \textbf{4.} randomly swap vowel columns within each group; \textbf{5.} randomly adjust the intensity of each vowel group by multiplying with a factor, which we use (0.5,2) in the implementation. \textbf{6.} denormalize the spectrogram.

After augmentation, the training set preserves the correct mapping between the spectrogram and transcription while enhancing the acoustic variations within the spectrogram. As seen through Figure \ref{fig:spec}, the overall correctness is preserved, but the details are now varied compared to the original spectrogram.

\subsection{Results}
\label{sec:results}
We present some preliminary results on pre-training the Whipser-base \cite{radford2022robustspeechrecognitionlargescale} architecture only using the Librispeech-960h dataset. For evaluation of its robustness, we evaluate the performance of the pre-trained model on Librispeech's test dataset and three out-of-distribution datasets: L2-Arctic \cite{zhao18b_interspeech} signifying accented speech, My Science Tutor (MyST) \cite{pradhan2023sciencetutormyst} signifying school-aged children speech, and ENNI \cite{liu2024fasaflexibleautomaticspeech}, a semi-proprietary dataset containing young children's speech that did not appear in OpenAI Whisper's pre-training.


As shown in Table \ref{main_table}, the pre-trained ASR model shows a higher generalization ability to out-of-distribution speech using acoustic augmentation strategies (Ours) compared to other variants.
While there is still a big gap to OpenAI's Whisper (Base-680k), acoustic augmentations achieve up to 19.24\% lower WER compared to the model trained without augmentation, 3.57\% compared with Mixup \cite{zhang2017mixup}, and 4.42\% compared to SpecMix \cite{kim21c_interspeech}. 
Moreover, acoustic augmentation has the potential to be applied with more training iterations, as shown in the continuing improvements between 20k and 60k training hours. 

\subsection{Synthetic Data with (Limited) Acoustic Variations}

Next, we hypothesize that adding acoustic variations can increase the acoustic space of the synthetic data, which can lead to better model robustness after pre-training. We conduct an ablation study by varying the acoustic features of speech samples produced by the TTS model. If acoustic diversity is indeed important, the pre-trained ASR model should be able to generalize more to real human speech with acoustic augmentation. Thus, we extend our experiments in Section \ref{sec:linguistic} by introducing random variations on the timbre, speed, pitch, and amplitude.

\noindent\textbf{Experiment Setup} Besides the acoustic augmentation strategies discussed in Section \ref{sec:details}, timbre variations are achieved by using the 54 voice packs provided with the Kokoro model. During voice synthesis, a randomly selected voice pack is used. 

By training on around 2-3\% of the corpus (around 2.5M samples), the ASR model reaches a stable position where the validation loss stops decreasing. At this moment, the WER on Librispeech is around 70-80\%, whereas the WER on the synthetic data remains close to 0. Further training brings WER on Librispeech back to greater than 100\%. While the WER is still not promising, experiment results indicate that acoustic variations can help the ASR model to pick up at least some words in the out-of-distribution speech sample.


%% file: table/main_table.tex
\begin{table*}[]
\tabcolsep 9.4pt
\caption{WER (\%) for pretraining the Whisper-base architecture with various data augmentation strategies. Training hours refer to the total hours of audio used in pre-training. All model checkpoints are picked using the lowest WER on the validation set of Librispeech.  }
\begin{tabular}{lcccccc}
\hline
                 &                      & \multicolumn{2}{c}{Librispeech 960h \cite{librispeech}}                  & \multicolumn{1}{c}{\multirow{2}{*}{ENNI \cite{liu2024fasaflexibleautomaticspeech}}} & \multicolumn{1}{c}{\multirow{2}{*}{MyST \cite{pradhan2023sciencetutormyst}}} & \multicolumn{1}{c}{\multirow{2}{*}{L2-arctic \cite{zhao18b_interspeech}}} \\
                 & Max Training Data (Hours)   & \multicolumn{1}{c}{clean} & \multicolumn{1}{c}{other} & \multicolumn{1}{c}{}                      & \multicolumn{1}{c}{}                      & \multicolumn{1}{c}{}                           \\ \hline
No Augmentation      & \multirow{6}{*}{20k} & 14.85                    & 29.00                      & 90.44                                    & 77.28                                    & 56.57                                         \\
SpecAugment \cite{park19e_interspeech}    &                      & 13.34                    & 24.41                    & 92.25                                    & 64.96                                    & 46.11                                         \\
Mixup \cite{zhang2017mixup}    &                      & 10.44                    & 23.55                    & 85.99                                    & 64.77                                    & 44.54                                         \\
SpecMix \cite{kim21c_interspeech}     &                      & \textbf{10.37}                    & \textbf{22.30}                     & 85.34                                    & 67.50                                     & \textbf{44.14}                                         \\
Ours &                      & 11.27                    & 25.01                    & \textbf{82.42}                                    & \textbf{63.08}                                    & 44.90                                          \\ \hline
Ours & \multirow{1}{*}{40k} & 10.11                  & 23.22                   & 84.16                                    & 61.66                                    & 43.44                                         \\\hline
Ours & \multirow{1}{*}{60k} & \textbf{8.58}                    & \textbf{21.84}                    & 82.67                                    & \textbf{58.04}                                    & \textbf{40.28}                                         \\
\hline
Base-680k   \cite{radford2022robustspeechrecognitionlargescale}     & 680k                 & 5.93                    & 13.00                      & 50.88                                    & 31.74                                    & 21.99                                         \\\hline
\end{tabular}
\label{main_table}
\end{table*}

%% file: table/pitch.tex
\begin{table}[h]
    \centering
    \tabcolsep 7pt
    \caption{Augmentation parameters for pitch.}
    \begin{tabular}{cccc}
    \hline
        Gender & Probability & Lower Bound & Upper Bound \\ \hline
        Male & 0.2 & -2 & 0 \\ 
        Male & 0.3 & 0 & 4 \\ 
        Female & 0.3 & -4 & 0 \\ 
        Female & 0.3 & 2 & 6 \\ \hline
    \end{tabular}
        \label{pitch_table}
\end{table}

%% file: 6_conclusion.tex
\section{Conclusion}
In conclusion, through analysis and experiments, this work reveals that robust ASR performance existing in foundation ASR models relies primarily on acoustic diversity as opposed to linguistic diversity in the training data. When data sources are limited, acoustic augmentations can significantly outperform existing data augmentation methods on unseen speech. We hope that our findings could offer insights into the creation of innovative pre-training data augmentation methods and foundation ASR models without requiring massive training data in the future. 
\clearpage